\pdfoutput=1

\documentclass[11pt]{article}
\usepackage{placeins}
\usepackage{algorithm}
\usepackage{algorithmic}
\usepackage{amsmath}
\usepackage{amssymb}
\usepackage{url}

\usepackage[preprint]{acl}

\usepackage{times}
\usepackage{latexsym}
\usepackage{booktabs}
\usepackage{microtype}

\usepackage[T1]{fontenc}

\usepackage[utf8]{inputenc}

\usepackage{microtype}

\usepackage{inconsolata}

\usepackage{graphicx}

\newcommand{\knnAHM}{\textbf{\text{knn}$_{\text{AHM}}$}}
\newcommand{\mahAHM}{\textbf{\text{mah}$_{\text{AHM}}$}}

\title{Out-of-Distribution Detection with Attention Head Masking for Multimodal Document Classification}

\author{
  Christos Constantinou$^{1,2,5,}$\thanks{Work does not relate to position at Amazon.},\  Georgios Ioannides$^{2,3,5,}$\footnotemark[1],\  Aman Chadha$^{2,4,5,}$\footnotemark[1]\ \\
  \textbf{Aaron Elkins}$^{5}$,\ \textbf{Edwin Simpson}$^{1}$ \\
  $^1$University of Bristol\quad
 $^2$Amazon GenAI\quad
 $^3$Carnegie Mellon University\\
  $^4$Stanford University\quad
 $^5$James Silberrad Brown Center for Artificial Intelligence\\
  \quad
  \texttt{christos.constantinou@bristol.ac.uk, gioannid@alumni.cmu.edu, hi@aman.ai,} \\
  \texttt{aelkins@sdsu.edu, edwin.simpson@bristol.ac.uk} \\
}

\begin{document}

\setlength{\textfloatsep}{10pt plus 1.0pt minus 2.0pt}
\maketitle
\begin{abstract}
Detecting out-of-distribution (OOD) data is crucial in machine learning applications to mitigate the risk of model overconfidence, thereby enhancing the reliability and safety of deployed systems. The majority of existing OOD detection methods predominantly address uni-modal inputs, such as images or texts. In the context of multi-modal documents, there is a notable lack of extensive research on the performance of these methods, which have primarily been developed with a focus on computer vision tasks. We propose a novel methodology termed as attention head masking (AHM) for multi-modal OOD tasks in document classification systems. Our empirical results demonstrate that the proposed AHM method outperforms all state-of-the-art approaches and significantly decreases the false positive rate (FPR) compared to existing solutions up to 7.5\%. 
This methodology generalizes well to multi-modal data, such as documents, where visual and textual information are modeled under the same Transformer architecture. To address the scarcity of high-quality publicly available document datasets and encourage further research on OOD detection for documents, we introduce FinanceDocs, a new document AI dataset. Our code\footnote{\url{https://github.com/constantinouchristos/OOD-AHM}} and dataset\footnote{\url{https://drive.google.com/drive/folders/1dV9obe_3hTsDoWJyYuNLBAXEiwOPwCw7}} are publicly available.

\end{abstract}

\begin{figure}[ht!]
    \centering
\includegraphics[width=0.88\columnwidth]{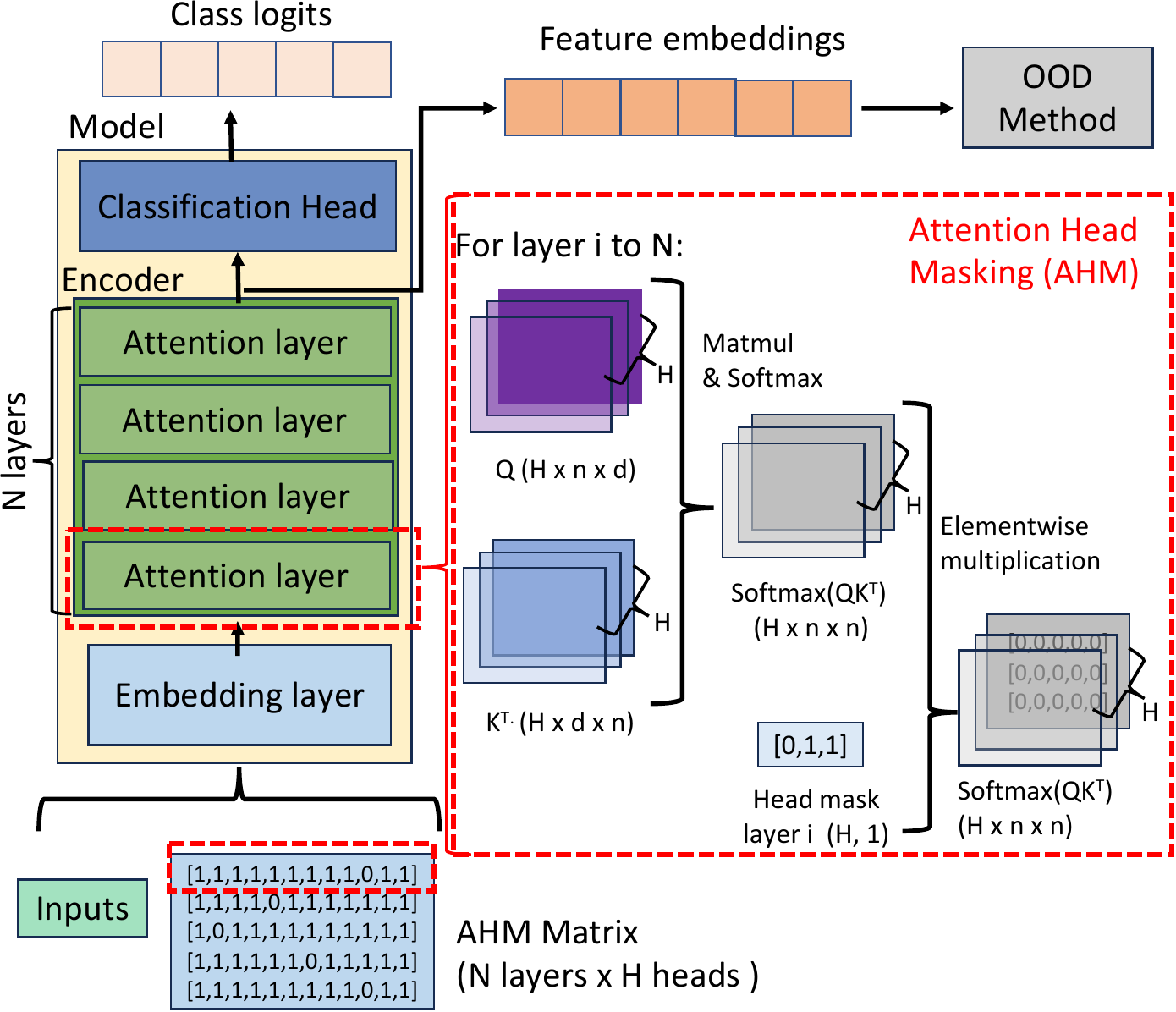} 
    \caption{Visual demonstration of AHM on a transformer-based model: For each attention layer, we utilize the corresponding attention head mask from the AHM matrix. Following query-key multiplication and the subsequent softmax operation, the resulting attention scores undergo element-wise multiplication with the relevant attention head mask. This process effectively reduces the attention scores of certain heads to zero, thereby inhibiting the propagation of their respective information through the value matrix.}
    \label{fig:figure1}
\end{figure}

\section{Introduction}

Out-of-distribution (OOD) detection presents a significant challenge in the field of document classification. When a classifier is deployed, it may encounter types of documents that were not included in the training dataset. This can lead to mishandling of such documents, causing additional complications in a production environment.

Effective OOD detection facilitates the identification of unfamiliar documents, enabling the system to manage them appropriately which allows the classifier to maintain its reliability and accuracy in real-world applications.

This has heightened the focus on OOD detection, where the primary objective is to determine if a new document belongs to a known in-distribution (ID) class or an OOD class. A significant challenge lies in the lack of supervisory signals from the unknown OOD data, which can encompass any content outside the ID classes. The complexity of this problem increases with the semantic similarity between the OOD and ID data \citep{fort2021exploring}.

\textls[-12]{A number of approaches have been developed to differentiate OOD data from ID data, broadly classified into three categories: (i) confidence-based methods, which focus on softmax confidence scores \citep{DBLP:journals/corr/abs-2010-03759, DBLP:journals/corr/HendrycksG16c, DBLP:journals/corr/abs-1911-11132, DBLP:journals/corr/abs-2110-00218, DBLP:journals/corr/LiangLS17}, (ii) features/logits-based methods, which emphasize logit outputs \citet{DBLP:journals/corr/abs-2111-09805, DBLP:journals/corr/abs-2111-12797, wang2022vim, djurisic2023extremely}, and (iii) distance/density-based methods, which concentrate on dense embeddings from the final layers \citep{ming2023exploit, lee2018simple, sun2022outofdistribution}. Recent research also investigates domain-invariant representations, such as HYPO \citep{ming2024hypo}, and introduces new OOD metrics like NECO \citep{ammar2024neco}, which leverage neural collapse properties \citep{Papyan_2020}. 
Confidence-based methods can be unreliable as they often yield overconfident scores for OOD data. Features/logits-based methods attempt to combine class-agnostic scores from the feature space with the ID class-dependent logits. Our approach focuses on identifying more robust class-agnostic scores from the feature space, and as such, we conduct our experiments using distance/density-based methods.}

Many OOD detection techniques have been developed, but most have been evaluated only in uni-modal systems, such as text or images, and not extensively tested in the document domain \citep{gu-etal-2023-critical}. This may be due to the lack of high-quality public document datasets, mostly based on IIT-CDIP \citep{Lewis2006}. To address the lack of comprehensive research in the document domain, we introduce a new document AI dataset, FinanceDocs. Additionally, we propose a novel technique called attention head masking (AHM) to effectively improve feature representations for distinguishing between ID and OOD data. Our method is illustrated in Figure~\ref{fig:figure1}. Our contributions can be summarized as follows:

\textbf{(1) FinanceDocs Dataset}: We introduce FinanceDocs, the first high-quality digital document dataset for OOD detection with multi-modal documents, offering digital PDFs instead of low-quality scans. \textbf{(2) AHM}: We propose a multi-head attention masking mechanism for transformer-based models applied post-fine-tuning. By identifying masks that enhance similarity between ID training and evaluation features, we generate robust representations that improve the separation of ID and OOD data using distance/density-based OOD techniques. Our AHM method surpasses existing OOD solutions on key metrics.

\section{Related Work}

Learning embedding representations that generalize effectively and facilitate better differentiation between ID and OOD data is a well-recognized challenge in the field of machine learning \citep{9847099}. To tackle this challenge, various studies have focused on specialized learning frameworks aimed at optimizing intra-class compactness and inter-class separation \citep{ye2021theoretical}. Building on the principles of contrastive representation learning, researchers such as \citet{pmlr-v119-chen20j} and \citet{li2021prototypical} introduced prototypical learning (PL). This approach leverages prototypes derived from offline clustering algorithms to enhance unsupervised representation learning. Furthermore, \citet{ming2024hypo} integrated PL into their OOD learning framework, HYPO, achieving effective separation between ID and OOD data. This line of research was further advanced by \citet{lu2024learning}, who introduced the concept of multiple prototypes per cluster and employed a maximum likelihood estimation (MLE) loss to ensure that sample embeddings closely align with their corresponding prototypes. Additionally, approaches such as VOS \citep{du2022vos} and NPOS \citep{tao2023nonparametric} have focused on regularizing the decision boundary between ID and OOD data by generating synthetic OOD samples, while \citet{lin2023flatsprincipledoutofdistributiondetection} utilized open-source data as an OOD signal.

In our proposed methodology, we similarly aim to enhance the distinction between ID and OOD data through improved embedding representations. However, unlike previous studies that explore customized learning frameworks diverging from the standard cross-entropy loss, we concentrate on feature regularization during inference using our proposed attention head masking methodology. Our approach deliberately avoids altering the network's training procedure, thereby mitigating potential negative impacts on performance and preventing increased training costs. By focusing on inference rather than training modifications, our method ensures robust and cost-effective OOD detection.

Other inference-based methods, such as Avg-Avg \citep{chen2022holisticsentenceembeddingsbetter} and Gnome \citep{chen2023finetuningdeterioratesgeneraltextual}, have also sought to enhance OOD detection through innovative techniques. Avg-Avg operates by averaging embeddings across both sequence length and different layers of a fine-tuned model, while Gnome combines embeddings from both a pre-trained and a fine-tuned model. These approaches, like our own, emphasize the importance of embedding manipulation during inference to achieve improved OOD detection without modifying the underlying training framework.

\section{Method}

The proposed AHM method, focuses on the feature extraction mechanisms inherent in transformer models, specifically the self-attention mechanism \cite{DBLP:journals/corr/VaswaniSPUJGKP17}. Based on the premise that OOD data exhibit less semantic similarity to ID data, our goal is to generate embedding features that enhance the separation between ID and OOD data. The embeddings are then used in distance or density-based OOD detection methods, such as the Mahalanobis \cite{lee2018simple} or kNN+ \cite{sun2022outofdistribution}. Our method is provided in Algorithm \ref{oodalgo} (cf. Appendix \ref{methodtheory} for the theoretical framework) and the masking step is summarised in Figure \ref{fig:figure1}. 

\begin{algorithm}[ht]
\caption{Optimization of Transformer-based Model using Attention Head Masking for OOD Detection -- cf. Appendix \ref{algot} for more details}
\begin{algorithmic}[1]
\STATE \textbf{Input:} Budget \( T \), model weights \( W_{\text{pretrained}} \), percentage masking  \( p \), neighbors \( K \), layers \( N \), attention heads \(H\), top attention head matrices to select \(F\)
\STATE \textbf{Output:} Optimal ensemble embedding

\STATE \textbf{1. Fine-tune Model} \( W_{\text{pretrained}} \) $\rightarrow$ \( W_{\text{finetuned}} \)

\FOR{trial = 1 to \( T \)}
    \STATE \textbf{2. InitializeAttention Head Matrix}
    \STATE Create \( N \times H \) matrix \( A \), \( A[i,j] = 1 \)

    \STATE \textbf{3. Mask Attention Heads}
    \STATE Randomly set elements of \( A[i,j] \) to 0

    \STATE \textbf{4. Extract Embeddings}
    \STATE Extract \( \text{embed}_{\text{train}} \in \mathbb{R}^{O \times Hid} \) and \( \text{embed}_{\text{eval}} \in \mathbb{R}^{Q \times Hid} \)

    \STATE \textbf{5. Compute Similarity Scores}
    \STATE For \( e_i \in \text{embed}_{\text{eval}} \), get \( K \) nearest neighbors in \( \text{embed}_{\text{train}} \) and compute mean score \( S_i \)

    \STATE \textbf{6. Assign and Collect Scores}
    \STATE Average similarity score: \( \frac{1}{Q} \sum_{i=1}^Q S_i \). Collect scores \(S_i \) and their respective \( A[i,j] \) 

\ENDFOR

\STATE \textbf{7. Select Top Scores}
\STATE Sort scores \(S_i \) , select top \( F \) masks  \( A[i,j] \)

\STATE \textbf{8. Ensemble Embedding Generation}
\STATE Use top \( F \) masks  \( A[i,j] \) to generate and average embeddings for OOD detection
\end{algorithmic}
\label{oodalgo}
\end{algorithm}

\begin{table*}[ht!]
\centering
\begin{minipage}{\textwidth} 
    \centering
    \footnotesize 
    \resizebox{\textwidth}{!}{ 
    \begin{tabular}{lcc|cc|cc|cc}
    \toprule
    \textbf{Method} & \multicolumn{2}{c}{\textbf{Tobacco3482 (ADVE OOD)}} & \multicolumn{2}{c}{\textbf{Tobacco3482 (Cross-dataset OOD)}} & \multicolumn{2}{c}{\textbf{FinanceDocs (Resume OOD)}} & \multicolumn{2}{c}{\textbf{FinanceDocs (Cross-dataset OOD)}} \\
    & \textbf{AUROC} & \textbf{FPR} & \textbf{AUROC} & \textbf{FPR} & \textbf{AUROC} & \textbf{FPR} & \textbf{AUROC} & \textbf{FPR} \\
    \midrule
    energy & 0.951 $\pm$ 0.012 & 0.267 $\pm$ 0.057 & 0.944 $\pm$ 0.014 & 0.157 $\pm$ 0.042 & 0.848 $\pm$ 0.093 & 0.413 $\pm$ 0.218 & 0.846 $\pm$ 0.016 & 0.567 $\pm$ 0.039 \\
    gradNorm & 0.940 $\pm$ 0.025 & 0.330 $\pm$ 0.116 & 0.824 $\pm$ 0.040 & 0.410 $\pm$ 0.094 & 0.742 $\pm$ 0.153 & 0.664 $\pm$ 0.251 & 0.724 $\pm$ 0.128 & 0.817 $\pm$ 0.145 \\
    kl & 0.914 $\pm$ 0.016 & 0.448 $\pm$ 0.099 & 0.970 $\pm$ 0.014 & 0.071 $\pm$ 0.035 & 0.902 $\pm$ 0.040 & 0.295 $\pm$ 0.106 & 0.840 $\pm$ 0.025 & 0.630 $\pm$ 0.047 \\
    knn & 0.958 $\pm$ 0.011 & 0.269 $\pm$ 0.074 & 0.991 $\pm$ 0.004 & 0.030 $\pm$ 0.018 & 0.965 $\pm$ 0.023 & 0.172 $\pm$ 0.127 & 0.891 $\pm$ 0.017 & 0.589 $\pm$ 0.067 \\
    Mahalanobis & 0.976 $\pm$ 0.009 & 0.155 $\pm$ 0.053 & 0.996 $\pm$ 0.002 & 0.010 $\pm$ 0.009 & 0.977 $\pm$ 0.013 & 0.122 $\pm$ 0.100 & 0.898 $\pm$ 0.017 & 0.541 $\pm$ 0.090 \\
    mah\textsubscript{AvgAvg} & 0.942 $\pm$ 0.008 & 0.375 $\pm$ 0.054 & 0.997 $\pm$ 0.001 & 0.0004 $\pm$ 0.0005 & 0.996 $\pm$ 0.003 & 0.006 $\pm$ 0.005 & 0.949 $\pm$ 0.015 & 0.353 $\pm$ 0.196 \\
    mah\textsubscript{Gnome} & 0.971 $\pm$ 0.009 & 0.155 $\pm$ 0.054 & 0.992 $\pm$ 0.003 & 0.037 $\pm$ 0.016 & 0.938 $\pm$ 0.035 & 0.314 $\pm$ 0.165 & 0.822 $\pm$ 0.024 & 0.646 $\pm$ 0.114 \\
    maxLogit & 0.946 $\pm$ 0.012 & 0.311 $\pm$ 0.063 & 0.945 $\pm$ 0.013 & 0.151 $\pm$ 0.033 & 0.851 $\pm$ 0.086 & 0.410 $\pm$ 0.203 & 0.846 $\pm$ 0.017 & 0.584 $\pm$ 0.037 \\
    msp & 0.929 $\pm$ 0.009 & 0.471 $\pm$ 0.103 & 0.952 $\pm$ 0.016 & 0.140 $\pm$ 0.050 & 0.883 $\pm$ 0.041 & 0.400 $\pm$ 0.142 & 0.846 $\pm$ 0.032 & 0.612 $\pm$ 0.048 \\
    neco & 0.971 $\pm$ 0.012 & 0.164 $\pm$ 0.046 & 0.995 $\pm$ 0.002 & 0.013 $\pm$ 0.011 & 0.975 $\pm$ 0.012 & 0.132 $\pm$ 0.096 & 0.888 $\pm$ 0.020 & 0.546 $\pm$ 0.114 \\
    residual & 0.976 $\pm$ 0.008 & 0.149 $\pm$ 0.051 & 0.996 $\pm$ 0.002 & 0.011 $\pm$ 0.009 & 0.976 $\pm$ 0.014 & 0.130 $\pm$ 0.106 & 0.896 $\pm$ 0.016 & 0.541 $\pm$ 0.089 \\
    vim & 0.976 $\pm$ 0.008 & 0.147 $\pm$ 0.044 & 0.996 $\pm$ 0.002 & 0.011 $\pm$ 0.009 & 0.976 $\pm$ 0.014 & 0.125 $\pm$ 0.101 & 0.899 $\pm$ 0.015 & 0.537 $\pm$ 0.086 \\
    \midrule
    {\knnAHM} & \textbf{0.969 $\pm$ 0.009} & \textbf{0.182 $\pm$ 0.039} & \textbf{0.991 $\pm$ 0.003} & \textbf{0.024 $\pm$ 0.013} & \textbf{0.975 $\pm$ 0.014} & \textbf{0.114 $\pm$ 0.088} & \textbf{0.885 $\pm$ 0.011} & \textbf{0.562 $\pm$ 0.096} \\
    {\mahAHM} & \textbf{0.985 $\pm$ 0.005} & \textbf{0.071 $\pm$ 0.041} & \textbf{0.997 $\pm$ 0.002} & \textbf{0.006 $\pm$ 0.006} & \textbf{0.978 $\pm$ 0.012} & \textbf{0.099 $\pm$ 0.086} & \textbf{0.892 $\pm$ 0.013} & \textbf{0.522 $\pm$ 0.126} \\    
    \textbf{mah\textsubscript{AvgAvg\_AHM}} & \textbf{0.956 $\pm$ 0.007} & \textbf{0.267 $\pm$ 0.007} & \textbf{0.998 $\pm$ 0.001} & \textbf{0.0001 $\pm$ 0.0009} & \textbf{0.996 $\pm$ 0.003} & \textbf{0.004 $\pm$ 0.003} & \textbf{0.951 $\pm$ 0.012} & \textbf{0.302 $\pm$ 0.012} \\    
    \bottomrule
    \end{tabular}
    }
\end{minipage}
\caption{Performance metrics (arithmetic mean and standard deviation) for different methods across two datasets with intra-dataset and cross-dataset experiments configurations per dataset using AUROC (higher is better) and FPR (lower is better) -- (cf. Appendix \ref{hyperparam} for hyperparameter tuning details).}
\vspace{-2mm}
\label{tab:combined_results}
\end{table*}
\section{Results and Discussion}

\subsection{Datasets}

We utilized two datasets in our experiments: Tobacco3482 and FinanceDocs. The Tobacco3482 dataset \citep{KUMAR2014119} comprises 10 classes: Memo (619), Email (593), Letter (565), Form (372), Report (261), Scientific (255), Note (189), News (169), Advertisement (162), and Resume (120). As a subset of IIT-CDIP \citep{Lewis2006}, it was further processed to remove blank and rotated pages, preserving the rich textual and image modalities essential for a multi-modal system. Despite these efforts, some instances exhibit poor OCR quality due to the low-quality scans.

\textls[-3]{We present FinanceDocs (cf. Appendix \ref{financedocs-desc} for per-category details and \ref{financedocs} for dataset samples), a newly created dataset comprising 10 classes derived from open-source financial documents, including SEC Form 13 (663), Financial Information (360), Resumes (287), Scientific AI Papers (267), Shareholder Letters (256), List of Directors (188), Company 10-K Forms (181), Articles of Association (176), SEC Letters (141), and SEC Forms (121). Unlike Tobacco3482, FinanceDocs consists of high-quality digital PDFs \citep{annualreports, secedgar, companieshouse, aclanthology, resumedataset}. The FinanceDocs dataset was labeled through the following process: a PDF parsing package (\texttt{PyPDF2}) was used to extract content from the original PDF documents. Each page was then visualized individually by a human annotator, who determined the relevance of the page to the collected classes and assigned the appropriate class label 
(cf. Appendix \ref{anno} for annotator training and validation).}

\vspace{-0.1in}
\subsection{Experimental Setup}

We employ two widely recognized OOD metrics to assess the performance of our proposed AHM method in comparison to other OOD benchmarks \citep{yang2024generalized}: AUROC, which measures the area under the ROC curve (higher values indicate better performance), and FPR, the false positive rate at a 95\% true positive rate. A higher AUROC signifies better discrimination, while a lower FPR indicates greater robustness in rejecting OOD data.

For our experiments, we utilize \texttt{LayoutLMv3} \citep{huang2022layoutlmv3}, a transformer-based multi-modal model with 125.92 million parameters. We conduct both cross-dataset and intra-dataset OOD experiments. In cross-dataset OOD, the model is trained on the classes of one dataset and evaluated on the entirety of the other dataset as OOD. In intra-dataset OOD, one of the 10 classes is designated as OOD, and the model is trained on the remaining 9 classes, with the ID data split into training and evaluation sets. We select Advertisement (ADVE) and Resumes as the OOD classes for Tobacco3482 and FinanceDocs, respectively.

The models are trained over 5 random runs, with checkpoints saved at high ID classification metrics. Checkpoints with low silhouette scores \(s(i) = \frac{b(i) - a(i)}{\max(a(i), b(i))}\) are filtered out to optimize intra-class similarity and inter-class separation. Our experiments were conducted using a single NVIDIA A100 GPU (80GB) for 72 GPU compute hours. We trained the models for a maximum of 15 epochs with an initial learning rate of 5$\times 10^{-5}$.

\subsection{Current Benchmarks}
We evaluated the peformance of various OOD detection methods, comparing them with our proposed methods, {\knnAHM}, {\mahAHM}, and \textbf{mah\textsubscript{AvgAvg\_AHM}}, which apply k-Nearest Neighbor (kNN) and Mahalanobis methods to dense embeddings generated by AHM. \textbf{mah\textsubscript{AvgAvg\_AHM}} is similar to {\mahAHM} but uses the AvgAvg embedding aggregation method \cite{chen2022holisticsentenceembeddingsbetter}.

\textls[-10]{As shown in Table \ref{tab:combined_results}, for the Tobacco3482 dataset with ADVE as the OOD class, our proposed {\mahAHM} outperformed other methods, achieving an AUROC of 0.985 and an FPR of 0.071. The high AUROC indicates that our method significantly enhances the Mahalanobis distance-based approach in distinguishing between ID and OOD samples. The notably lower FPR compared to previous methods like \textit{vim} and \textit{residual} (FPRs of 0.147 and 0.149, respectively) demonstrates the robustness of {\mahAHM} in correctly rejecting OOD samples.}

For the FinanceDocs dataset, with Resumes as the OOD class, both {\knnAHM} and {\mahAHM} achieved superior performance, with AUROCs of 0.975 and 0.978, and FPRs of 0.114 and 0.099, respectively. Our \textbf{mah\textsubscript{AvgAvg\_AHM}} method also improved performance over mah\textsubscript{AvgAvg}, highlighting the effectiveness of our approach in creating more separable embeddings between ID and OOD data. This is further evidenced by cross-dataset results in Table \ref{tab:combined_results}, where \textbf{mah\textsubscript{AvgAvg\_AHM}} consistently outperformed mah\textsubscript{AvgAvg}, notably reducing the FPR by 5\% on FinanceDocs and achieving an AUROC of 0.99 with an FPR of 0.0001 on Tobacco3482. This performance surpasses the respective method mah\textsubscript{AvgAvg} without AHM applied. In fact, across all methods tested mah\textsubscript{AvgAvg}, Mahalanobis and knn, the application of our AHM technique consistently resulted in improved performance.

\textls[0]{Overall, the AHM technique significantly enhances the performance of kNN, Mahalanobis, and mah\textsubscript{AvgAvg}, resulting in superior outcomes for {\knnAHM}, {\mahAHM}, and \textbf{mah\textsubscript{AvgAvg\_AHM}}, as evidenced by higher AUROCs and lower FPRs across intra-dataset and cross-dataset experiments, demonstrating strong generalizability across diverse datasets and methods.}

\section{Conclusion}

In this study, we present the AHM technique for OOD detection in transformer-based document classification. Our methods, {\knnAHM}, {\mahAHM} and \textbf{mah\textsubscript{AvgAvg\_AHM}}, demonstrated significant improvements in AUROC and FPR metrics across various datasets. These results underscore the effectiveness of optimizing attention mechanisms to enhance feature separation between ID and OOD data. Additionally, we introduce the FinanceDocs dataset, contributing valuable resources to OOD detection research. Our findings highlight AHM as a promising approach for achieving robust and accurate OOD detection in document classification.

\section{Limitations}
While AHM techniques significantly reduced FPR in most cases, the improvements were marginal in cross-dataset scenarios where the Tobacco dataset served as the OOD data. This suggests a potential dependency on specific datasets. Additionally, AHM is a technique limited to attention-based DNN architectures that employ multi-head self-attention. Future research should aim to broaden the range of datasets explored.

\bibstyle{IEEEtran}
\bibliography{refs.bib}

\pagebreak 
\clearpage
\appendix

\section{Appendix}

This section provides supplementary material in the form of dataset examples, implementation details, etc. to bolster the reader’s understanding of the concepts presented in this work.

\subsection{Proposed Methodology and Theoretical Framework}
\label{methodtheory}
The central hypothesis underlying the proposed solution is predicated on the assumption that ID data should exhibit greater similarity in their feature representations when compared to OOD data. Consequently, we posit that when considering a pair of data points from two similar ID classes (denoted as Pair A) and a pair consisting of one ID and one OOD data point (denoted as Pair B), the application of a masking procedure on input features (whether textual or visual) would result in a more pronounced divergence in the feature space for Pair B as compared to Pair A.
Initial experiments were conducted with random masking of input features. For textual data, this involved replacing tokens randomly with the `\texttt{[MASK]}' token. For visual data, random image patches were set to zero, effectively splitting the image into patches and nullifying selected segments. These preliminary experiments revealed two critical factors influencing the final feature embeddings used in distance-based OOD detection methods, such as the Mahalanobis distance: (a) the input tensors provided to the model, and (b) the feature extraction mechanism employed by the model, specifically the attention mechanism.

Although the early experiments primarily focused on input masking, achieving a consistent masking strategy proved challenging. While a consistent mask could be established for visual data by dividing images into uniformly sized chunks and consistently masking specific segments, such consistency was elusive for textual features. The variability in sequence length across different tokens complicated the masking process, often leading to strategies that involved masking padding tokens rather than meaningful data.

In light of these challenges, our focus shifted from input masking to the feature extraction process itself, particularly the attention mechanism within the model. We discovered that consistent masking could be achieved by selectively masking attention heads within different layers of the encoder. These heads are responsible for learning different representations and capturing different aspects of the input sequence. Hence by shutting down heads we are effectively deactivating certain pattern-extracting mechanisms within the attention architecture.

\subsection{Description of Algorithm 1}
\label{algot}
As detailed in Algorithm 1, we begin with a fine-tuned model and proceed by randomly initializing various attention head masks based on a masking hyperparameter \( p \). This hyperparameter represents the percentage of attention heads \( H \) set to zero within each attention layer \( N \) of the model. For each random mask, we extract dense hidden representations from both the training and evaluation datasets. The objective is to identify which of these randomly generated attention head masks minimizes the divergence between the representations of the evaluation and training data in the feature space. This is accomplished by calculating the average similarity score among the top \( K \) nearest neighbors for each evaluation data point.

The attention head masks are then ranked based on these aggregated similarity scores. Finally, we select the top \( F \) masks with the highest similarity scores between the evaluation and training data and use them to generate new feature representations. These features are then ensembled (i.e., averaged) and subsequently utilized in a distance-based OOD detection method, such as the Mahalanobis distance.

\subsection{Hyperparameter Tuning}
\label{hyperparam}
Table \ref{tab:hyperparameters} summarizes the hyperparameters for model training. The model was trained using a carefully selected set of hyperparameters to optimize its performance. The training batch size per device was set to 32, while the evaluation batch size was configured at 8, ensuring efficient computation throughout the process. To stabilize updates, gradient accumulation was performed over 8 steps. The learning rate was set at \(5 \times 10^{-5}\), with no weight decay applied, to prevent the risk of overfitting.

The Adam optimizer was configured with parameters \(\beta_1 = 0.9\), \(\beta_2 = 0.999\), and an epsilon value of \(1 \times 10^{-8}\) to ensure effective convergence. To maintain stability during training, the maximum gradient norm was capped at 1.0. The model underwent training for 65 epochs, with evaluations delayed by 5 steps to monitor progress at appropriate intervals, allowing for a well-tuned and stable learning process.

The hyperparameters chosen for the proposed AHM method are presented in Table \ref{tab:hyperparametersAHM}. Following the procedure outlined in Algorithm 1, an exploration budget of 25 was allocated for potential AHM configurations. To assess the effectiveness of different configurations, masking percentages of 0.1 and 0.2 were applied during the process.

To ensure robust performance, similarity scores between ID validation data and ID training data were computed. These scores were determined by averaging the similarity of the top 10 nearest neighbors for each validation data point. Using these similarity scores, the top five AHM heads were selected to generate the final representation embeddings, which were then combined through an ensemble approach to enhance the overall model performance.

\begin{table}[h!]
    \centering
    \caption{Hyperparameters for model training.}
    \begin{tabular}{lc}
        \toprule
        \textbf{Hyperparameter} & \textbf{Value} \\
        \toprule
        per\_device\_train\_batch\_size & 32 \\
        per\_device\_eval\_batch\_size & 8 \\
        gradient\_accumulation\_steps & 8 \\
        eval\_delay & 5 \\
        learning\_rate & 5e-05 \\
        weight\_decay & 0.0 \\
        adam\_beta1 & 0.9 \\
        adam\_beta2 & 0.999 \\
        adam\_epsilon & 1e-08 \\
        max\_grad\_norm & 1.0 \\
        num\_train\_epochs & 65 \\
        \bottomrule
    \end{tabular}
    \label{tab:hyperparameters}
\end{table}

\begin{table}[h!]
    \centering
    \caption{Hyperparameters for AHM.}
    \begin{tabular}{lc}
        \toprule        
        \textbf{Hyperparameter} & \textbf{Value} \\
        \toprule        
        Exploration budget ($T$) & 25 \\
        Percentage masking ($p$) & [0.1, 0.2] \\
        Neighbors ($K$) & 10 \\
        Top AHM matrices select ($F$) & 5 \\
        \bottomrule
    \end{tabular}
    \label{tab:hyperparametersAHM}
\end{table}

\subsection{Annotator Training and Validation}\label{anno}
To maintain high-quality annotation in line with ethical standards, we enlisted three postgraduate students fluent in English. They received instruction and participated in sessions with finance professionals to address any task-related questions. The annotation process spanned about four months, involving 90 training sessions, with breaks scheduled every 45 minutes. The students were compensated through gift vouchers and honorariums per minimum wage requirements\footnote{\url{https://www.minimum-wage.org/international/united-states}}.

\subsection{Dataset description of FinanceDocs}\label{financedocs-desc}

The FinanceDocs dataset comprises a diverse collection of financial and legal documents sourced from various reliable platforms, offering a comprehensive view of corporate disclosures, shareholder communications, and regulatory filings. Each document type serves a distinct purpose, providing insights into different aspects of corporate governance, financial performance, and regulatory compliance, as detailed below:

\begin{itemize}
\item \textbf{SEC form documents:} These documents were collected from the Securities Exchange Commission (SEC) website. These forms are statements of changes in beneficial ownership.

\item \textbf{Shareholder letter documents}: These documents were collected from annual reports. A shareholder letter in an annual report provides a summary of the company’s financial performance, highlighting key achievements, strategic initiatives, and market conditions over the past year. It offers leadership's perspective on successes and challenges while outlining future goals and potential risks. The letter also emphasizes the company’s commitment to corporate governance, social responsibility, and long-term growth.

\item \textbf{SEC letter documents:} These documents were collected from the SEC website. These are letters from companies to the SEC about various company disclosures.

\item \textbf{SEC-13 form documents:} These documents were collected from the SEC website. These forms disclose significant information about an entity's ownership or control over securities, typically required for investors with large holdings.

\item \textbf{10k form documents:}
These documents were collected from annual reports. These represent the 10k forms of an annual report
 
\item \textbf{Financial info documents:}
These documents were collected from annualreports \citep{annualreports}. They consist of various financial information, including the income statement, balance sheet, and cash flow statement, which detail the company’s revenue, expenses, assets, liabilities, and cash movements. It also includes financial ratios and metrics to assess profitability, liquidity, and leverage.

\item \textbf{Articles of scientific paper documents:} These documents were collected from ACL Anthology\footnote{\url{https://aclanthology.org/}}. It is a comprehensive digital archive of research papers in computational linguistics and natural language processing, published by the Association for Computational Linguistics.

\item \textbf{Articles of resume documents:} These documents were collected from Kaggle. They represent resumes from different occupations.

\item \textbf{Articles of Association documents:} These documents were collected from Companies House Services UK. They represent documents relating to articles of association of a company. These involve information such as directors powers and responsibilities, interpretation and limitation of liability as well as distribution of shares.

\item \textbf{Director documents:}
These documents were collected from annual reports and Companies House Services UK\footnote{\url{https://www.gov.uk/government/organisations/companies-house}}. It involves information about the directors of a company. 
 
\end{itemize} 

\onecolumn
\subsection{Dataset examples of FinanceDocs}
\label{financedocs}

Presented below are examples from each document category included in FinanceDocs, providing the reader with a comprehensive visual overview of the dataset.

\begin{figure}[htbp!]  
    \centering
    \includegraphics[width=0.95\textwidth]{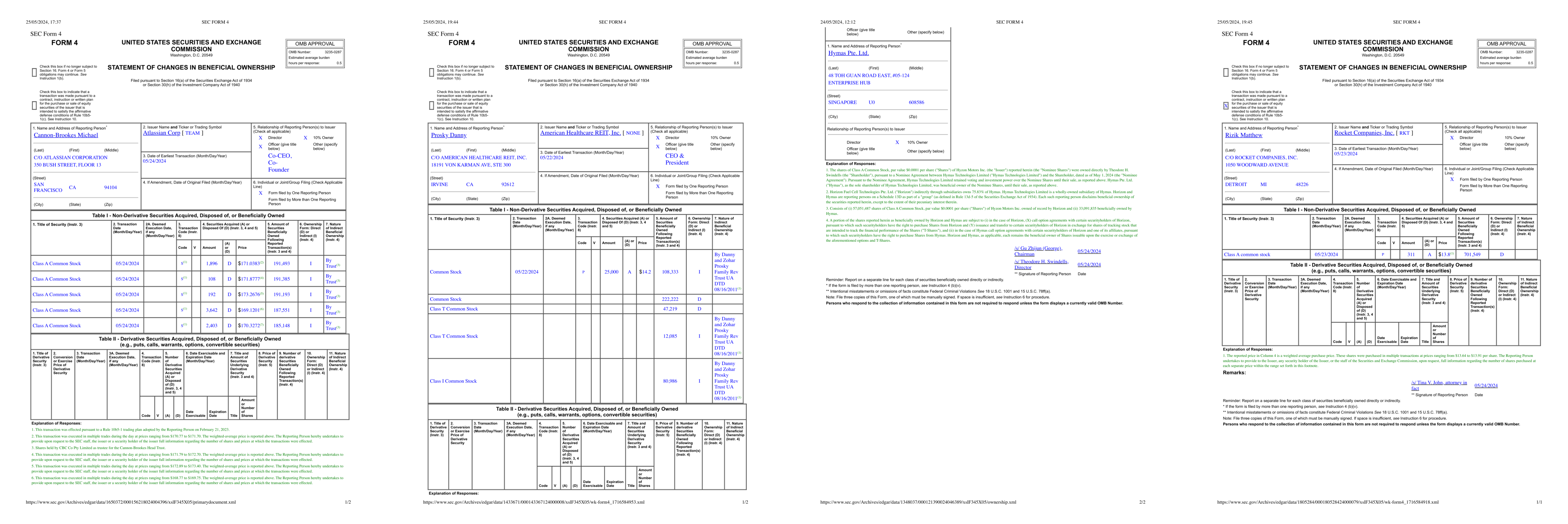}  
    \caption{Examples of SEC form documents.}
    \label{fig:sec_form}
\end{figure}

\vspace{0.5in}
 \begin{figure}[htbp!]  
    \centering
    \includegraphics[width=1\textwidth]{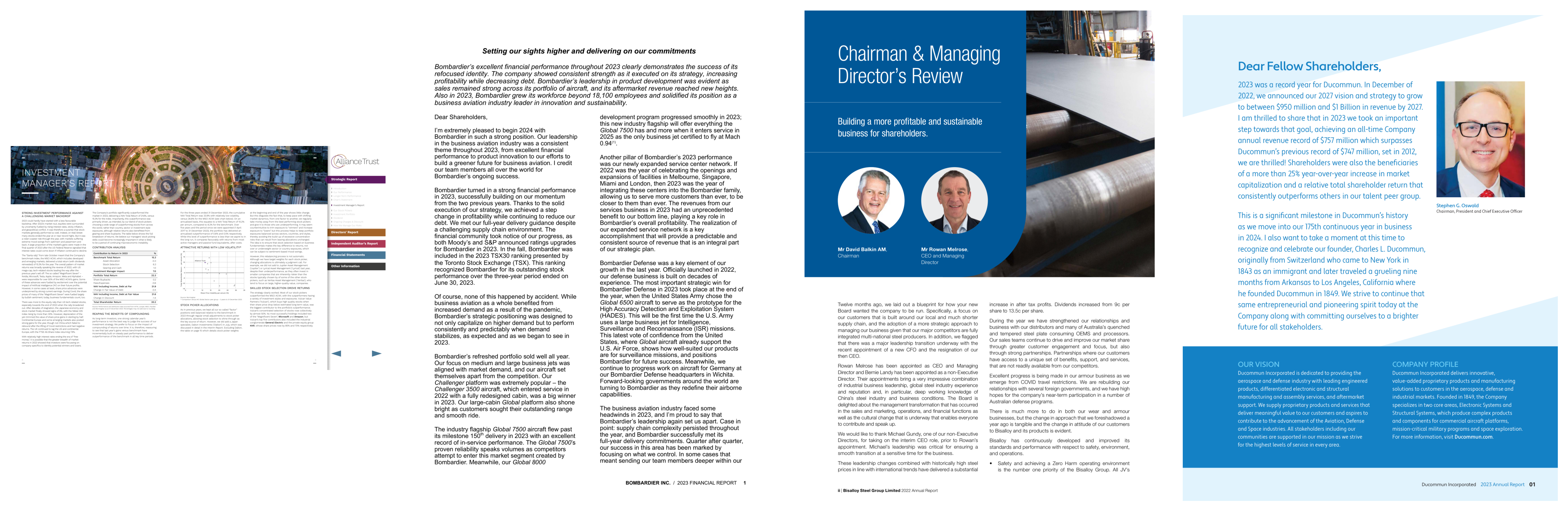}  
    \caption{Examples of shareholder letter documents.}
    \label{fig:shareholder_letter}
\end{figure}
\vspace{0.5in}

 \begin{figure}[htbp!]
     \centering
     \includegraphics[width=\textwidth]{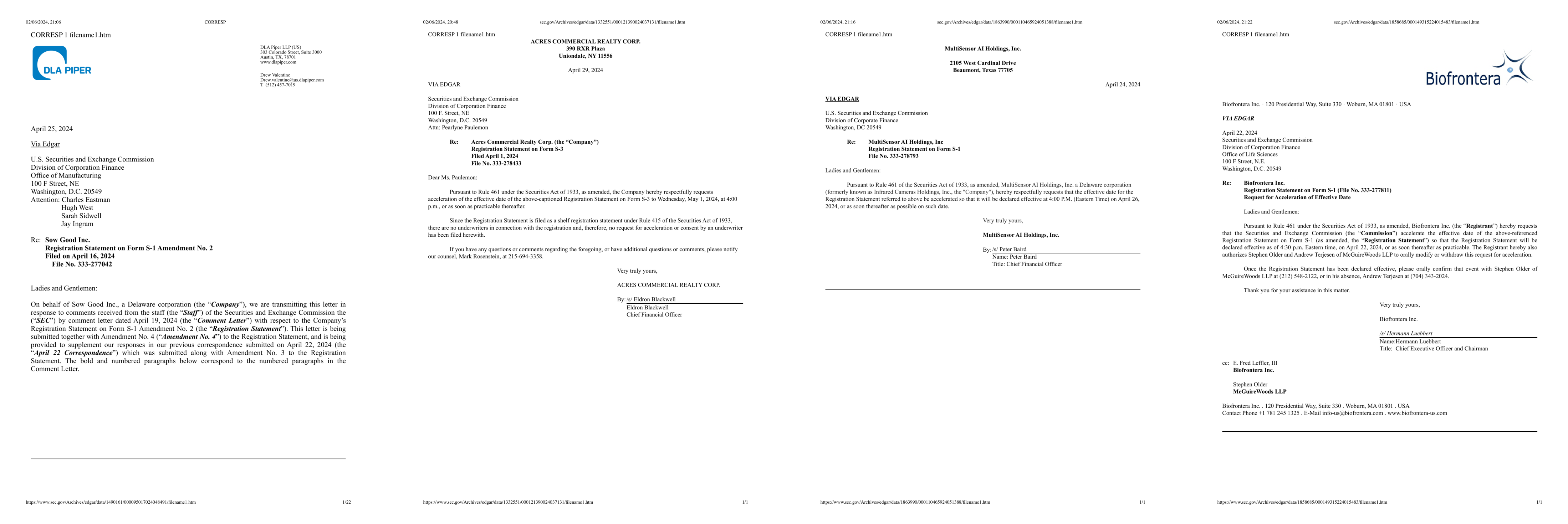}
     \caption{Examples of SEC letter documents.}
     \label{fig:sec_letter}
 \end{figure}

 \begin{figure}[htbp!]
     \centering
     \includegraphics[width=\textwidth]{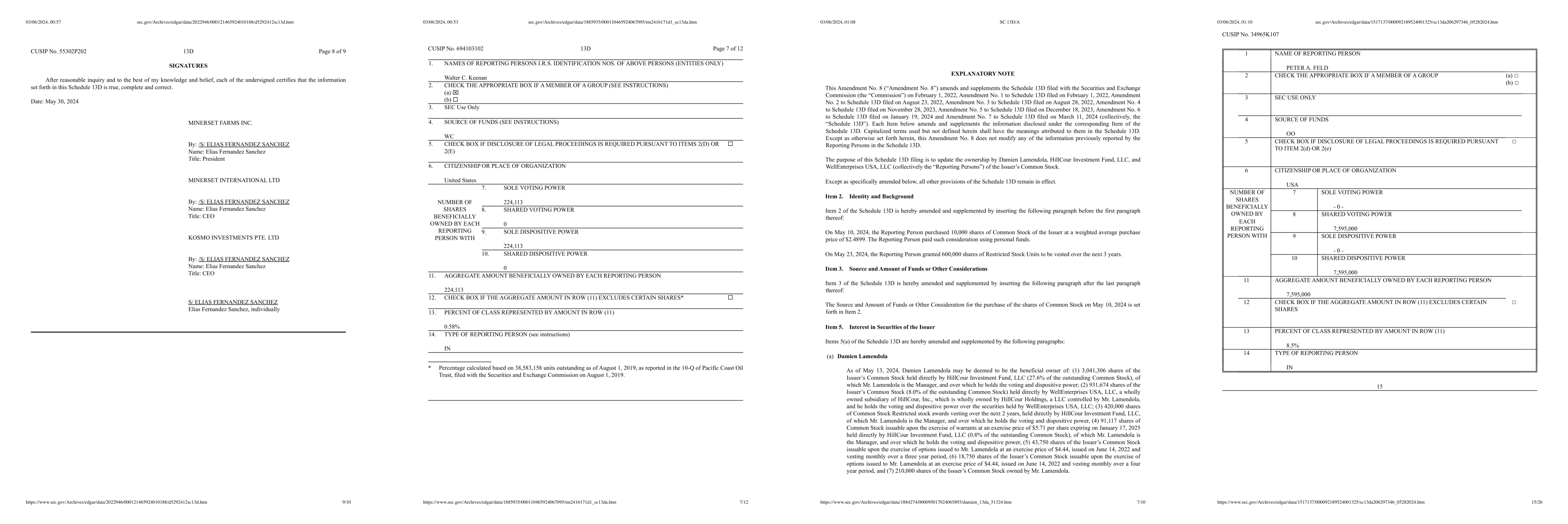}
     \caption{Examples of SEC-13 form documents.}
     \label{fig:sec_13}
 \end{figure}

 \begin{figure}[htbp!]
     \centering
     \includegraphics[width=\textwidth]{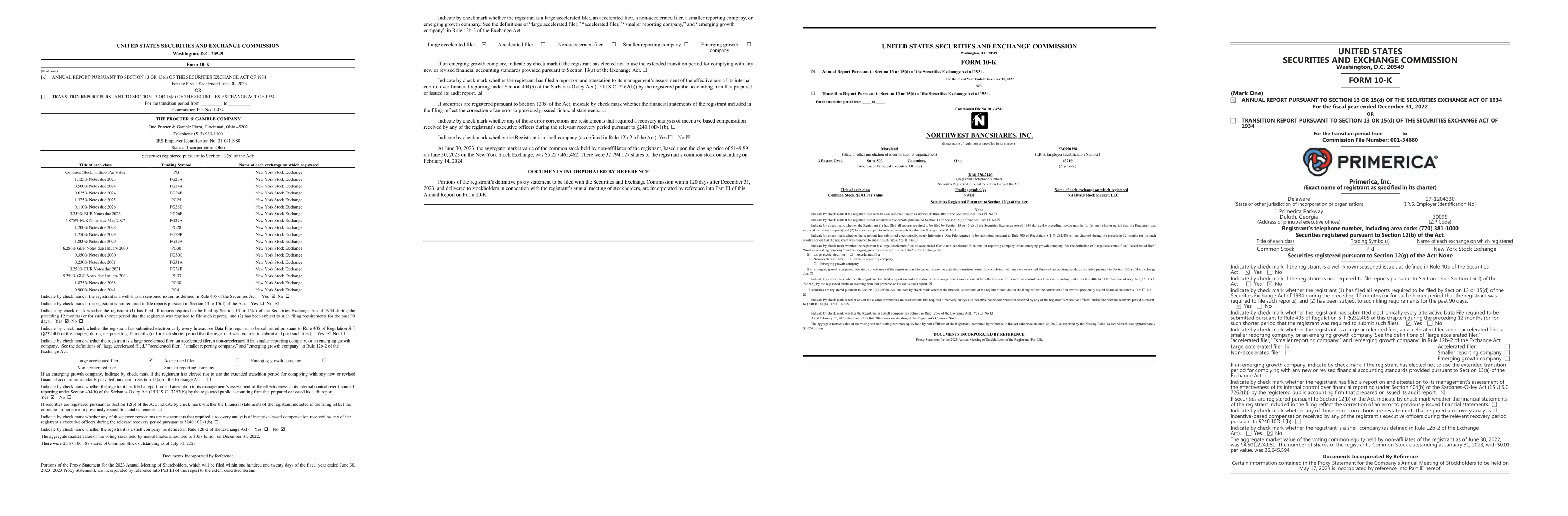}
     \caption{Examples of 10k form documents.}
     \label{fig:images__10k_page}
 \end{figure}

 \begin{figure}[htbp!]
     \centering
     \includegraphics[width=\textwidth]{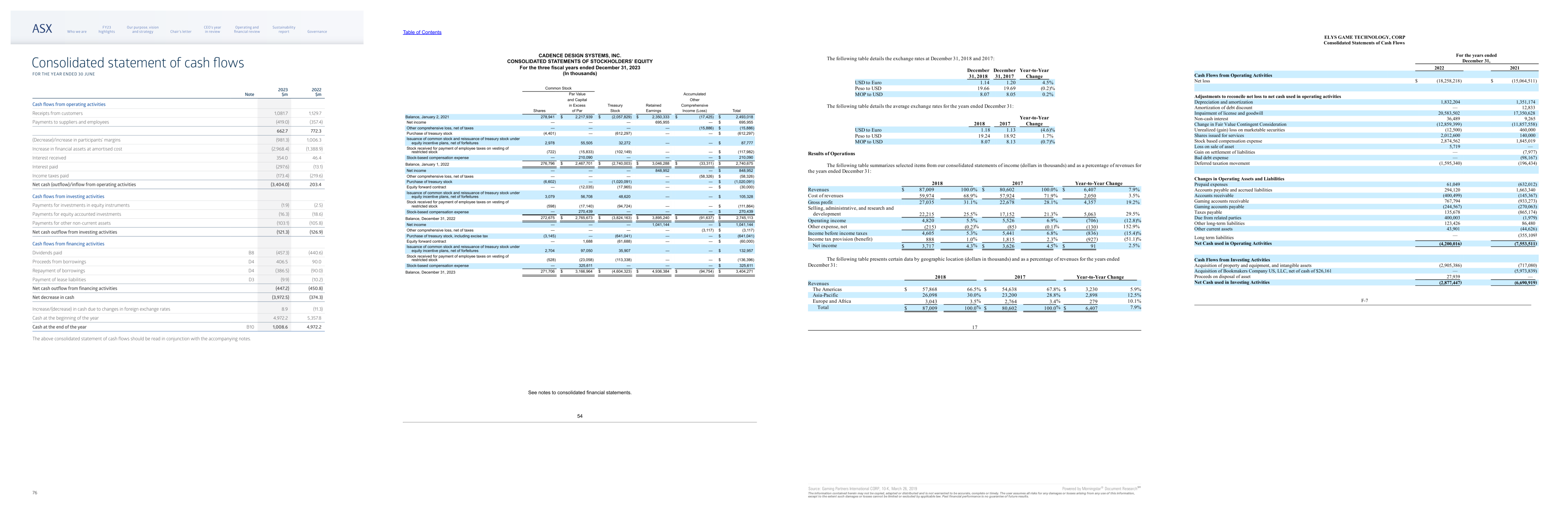}
     \caption{Examples of financial info documents.}
     \label{fig:financial_info}
 \end{figure}

 \begin{figure}[htbp!]
     \centering
     \includegraphics[width=\textwidth]{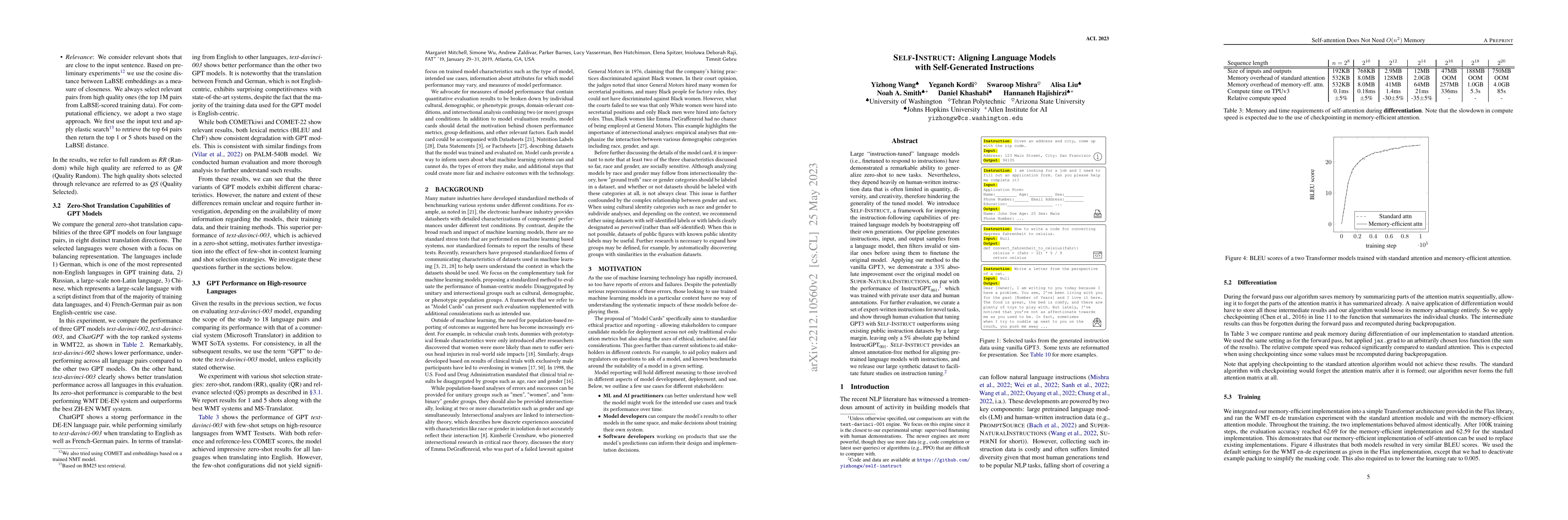}
     \caption{Examples of scientific paper documents.}
     \label{fig:scientific_ai_paper}
 \end{figure}

 \begin{figure}[htbp!]
     \centering
     \includegraphics[width=\textwidth]{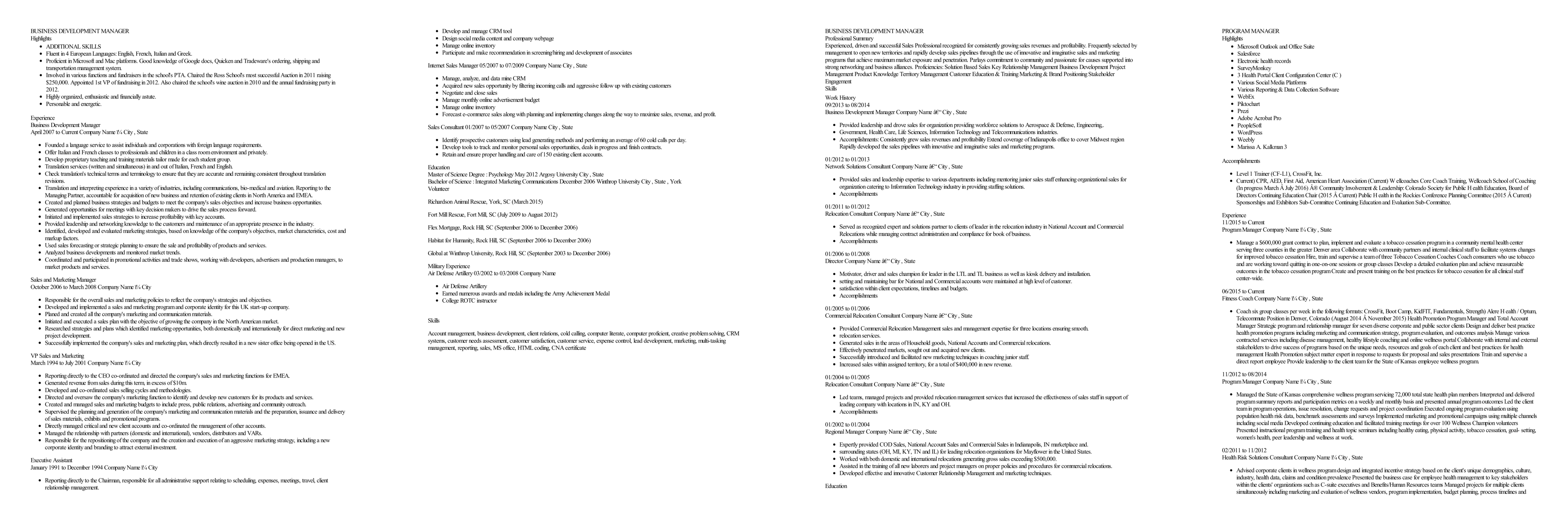}
     \caption{Examples of resume documents.}
     \label{fig:resumes}
 \end{figure}

 \begin{figure}[htbp!]
     \centering
     \includegraphics[width=\textwidth]{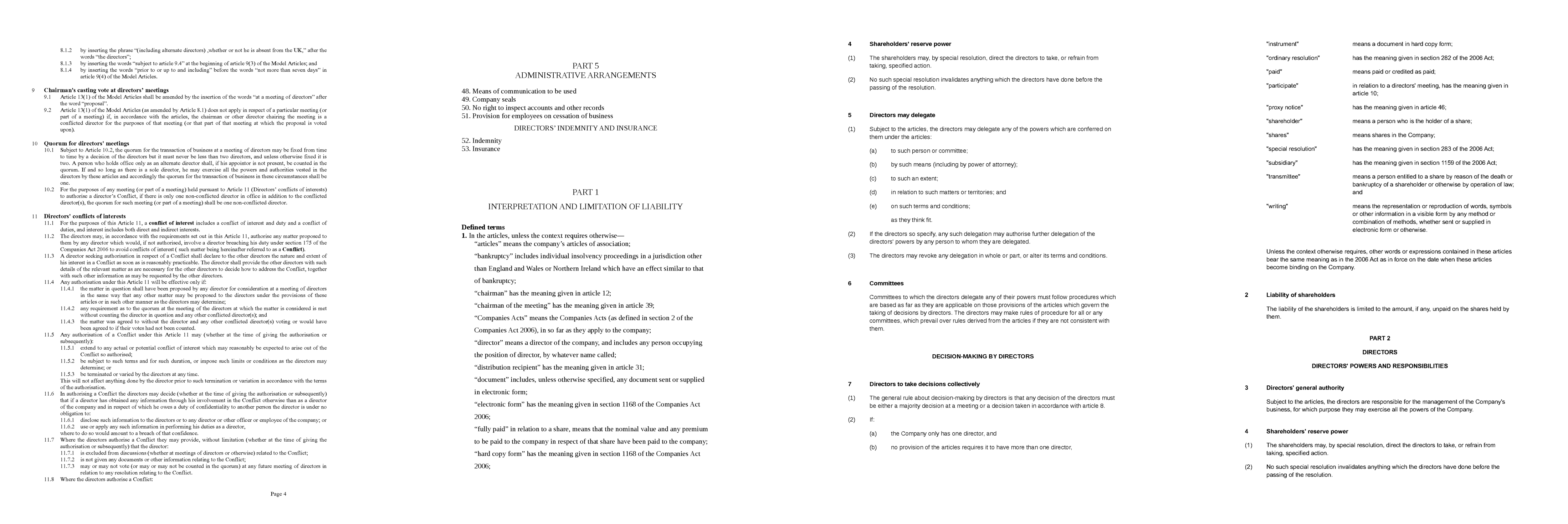}
     \caption{Examples of Articles of Association documents.}
     \label{fig:articles_of_association}
 \end{figure}

 \begin{figure}[htbp!]
     \centering
     \includegraphics[width=\textwidth]{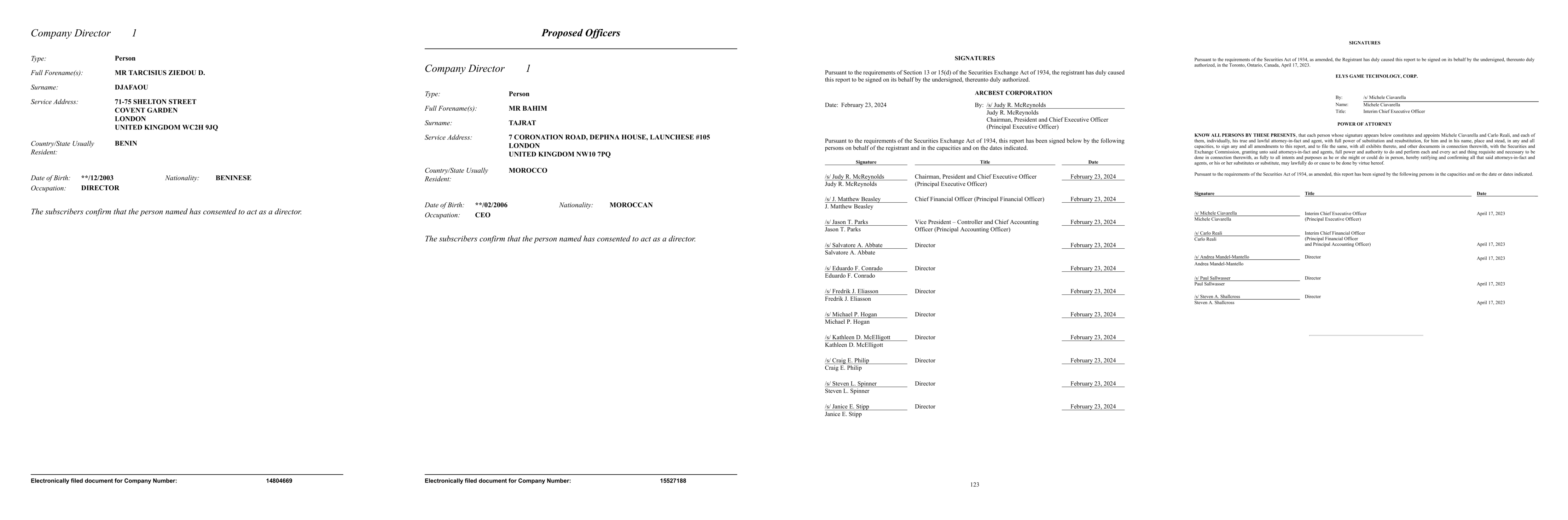}
     \caption{Examples of list of director documents.}
     \label{fig:list_directors}
 \end{figure}

\end{document}